\documentclass[sigconf]{acmart}
\AtBeginDocument{%
  \providecommand\BibTeX{{%
    \normalfont B\kern-0.5em{\scshape i\kern-0.25em b}\kern-0.8em\TeX}}}

\setcopyright{acmcopyright}
\copyrightyear{2023}
\acmYear{2023}
\setcopyright{acmlicensed}
\acmConference[SIGIR '23] {Proceedings of the 46th International ACM SIGIR Conference on Research and Development in Information Retrieval}{July 23--27, 2023}{Taipei, Taiwan.}
\acmBooktitle{Proceedings of the 46th International ACM SIGIR Conference on Research and Development in Information Retrieval (SIGIR '23), July 23--27, 2023, Taipei, Taiwan}
\acmPrice{15.00}
\acmISBN{978-1-4503-9408-6/23/07}
\acmDOI{10.1145/3539618.3592012}
\begin{document}

\title{LogicRec: Recommendation with Users' Logical Requirements}


\author{Zhenwei Tang}
\affiliation{%
  \institution{University of Toronto}
  \city{Toronto}
  \state{ON}
  \country{Canada}
}
\email{josephtang@cs.toronto.edu}

\author{Griffin	Floto}
\affiliation{%
  \institution{University of Toronto}
  \city{Toronto}
  \state{ON}
  \country{Canada}
}
\email{griffin.floto@mail.utoronto.ca}

\author{Armin Toroghi}
\affiliation{%
  \institution{University of Toronto}
  \city{Toronto}
  \state{ON}
  \country{Canada}
}
\email{armin.toroghi@mail.utoronto.ca}

\author{Shichao Pei}
\affiliation{%
  \institution{University of Notre Dame}
  \city{South Bend}
  \state{IN}
  \country{USA}
}
\email{spei2@nd.edu}

\author{Xiangliang Zhang}
\affiliation{%
  \institution{University of Notre Dame}
  \city{South Bend}
  \state{IN}
  \country{USA}
}
\email{xzhang33@nd.edu}

\author{Scott Sanner}
\authornote{Corresponding Author.}
\affiliation{%
  \institution{University of Toronto}
  \city{Toronto}
  \state{ON}
  \country{Canada}
}
\email{ssanner@mie.utoronto.ca}
\renewcommand{\shortauthors}{Tang, et al.}

\begin{abstract}
Users may demand recommendations with highly personalized requirements involving logical operations, e.g., the intersection of two requirements, where such requirements naturally form structured logical queries on knowledge graphs (KGs). To date, existing recommender systems lack the capability to tackle users' complex logical requirements. In this work, we formulate the problem of \underline{rec}ommendation with users' \underline{logic}al requirements (LogicRec) and construct benchmark datasets for LogicRec. Furthermore, we propose an initial solution for LogicRec based on \textit{logical requirement} retrieval and \textit{user preference} retrieval, where we face two challenges. First, KGs are incomplete in nature. Therefore, there are always missing true facts, which entails that the answers to logical requirements can not be completely found in KGs. In this case, item selection based on the answers to logical queries is not applicable. We thus resort to logical query embedding (LQE) to jointly infer missing facts and retrieve items based on logical requirements. Second, answer sets are under-exploited. Existing LQE methods can only deal with query-answer pairs, where queries in our case are the intersected user preferences and logical requirements. However, the logical requirements and user preferences have different answer sets, offering us richer knowledge about the requirements and preferences by providing requirement-item and preference-item pairs. Thus, we design a multi-task knowledge-sharing mechanism to exploit these answer sets collectively. Extensive experimental results demonstrate the significance of the LogicRec task and the effectiveness of our proposed method.
\end{abstract}

\ccsdesc[500]{Information systems~Query representation}
\ccsdesc[500]{Information systems~Recommender systems}
\ccsdesc[300]{Theory of computation~Logic}

\keywords{recommender system; knowledge graph; logical query answering}



\maketitle

\begin{figure}[t!]
	\centering  
	\includegraphics[width=0.48\textwidth]{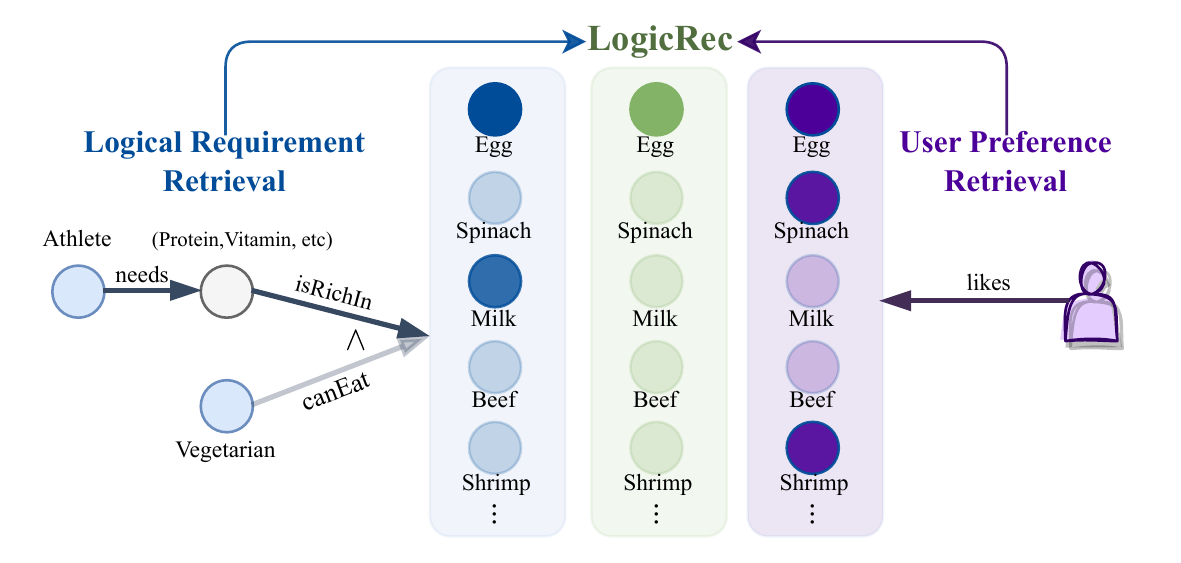}
	\caption{LogicRec is based on logical requirement retrieval and user preference retrieval. The shown user may just be finished working out and become a vegetarian recently. Therefore, the user requires food that vegetarians can eat and ($\wedge$) is rich in nutrition (the gray node) that athletes need, which forms the logical requirement.
 Note that there may be missing true facts in KGs, e.g, vegetarians can eat spinach (denoted by the light-colored arrow), yielding incomplete answer sets. Also, requirement retrieval and preference retrieval have different answer sets to be explored.
 }
	\label{fig:example}
\end{figure}

\section{Introduction}
Recommender systems are playing crucial roles in web and mobile applications, attracting a vast amount of interest from the industry and the academia~\cite{sedhain2015autorec,he2017neural,wang2019neural,ying2018sequential,guo2020survey}. Typically, user preferences are modeled from their interaction history for recommendation~\cite{zhao2021recbole,ko2022survey}. However, other than user preferences, users may have different needs each time they interact with a recommender system, which can be complex logical requirements. For example, as shown in Figure \ref{fig:example}, the user can ask the system to recommend food that not only vegetarians can eat but also is rich in nutrition that athletes need. 
Such logical requirements are essential to provide highly personalized recommendations.
In this case, the recommender system should be able to give recommendations based on both \textbf{user preferences} and \textbf{logical requirements}. 

Although efforts have been made in language-based recommender systems~\cite{zhou2020leveraging,zhou2020towards,zhou2020improving,li2022customized,li2022self,abdollah2023self} and critiquing-based recommender systems~\cite{luo2020deep, li2020ranking,yang2021bayesian,shen2022distributional} to jointly model users preference and their requirement, they typically only consider simple requirements without logical operations. Therefore, they can not fulfill users' needs for highly personalized recommendations with complex logical requirements. In this work, we focus on \underline{rec}ommendation with users' \underline{logic}al requirements (LogicRec) considering both the user preferences and the logical requirements, where the logical requirements are naturally formulated as structured logical queries over knowledge graphs (KGs). An example is shown in Figure \ref{fig:example}.

To approach LogicRec, the \textbf{KG incompleteness issue}. It is well-known that even the state-of-the-art KGs still suffer from severe incompleteness~\cite{rossi2021knowledge,Tang2022PositiveUnlabeledLW}, i.e., many true facts are missing. For example, more than 66\% of person entries in Freebase and DBpedia lack their birthplaces~\cite{dong2014knowledge,krompass2015type}. Therefore, the answer set to the logical requirements in LogicRec can not be completely found in KGs. Admittedly, an intuitive way to jointly consider logical requirements and user preferences is to select the items satisfying the logical requirements from the items users may like. However, under such a severe KG incompleteness situation, such a solution is not applicable.
For example, as shown in Figure \ref{fig:example}, the fact ``vegetarians can eat spinach'' is missing from the KG, thus the retrieved item set of the logical requirement will miss ``spinach''. To this end, we resort to logical query embedding (LQE) methods \cite{hamilton2018embedding,ren2019query2box,ren2020beta} to jointly infer the missing facts in the KG and retrieve items based on the logical requirements and user preferences.
On the other hand, we face the \textbf{answer set under-exploitation issue}.
Existing LQE methods can only deal with query-answer pairs, where a \textit{LogicRec query} in our case is the \textit{intersected user preference query and logical requirement query}. However, as shown in Figure \ref{fig:example}, the logical requirement query and user preference query have different answer sets (in blue and purple), offering us richer knowledge about the requirements and preferences by providing requirement-item and preference-item pairs. If we directly apply LQE methods for the query-answer pairs, only the answer set in green shown in Figure \ref{fig:example} would be accessible. Thus, we design a multi-task knowledge-sharing framework to collectively exploit the answer sets of the user preference queries, the logical requirement queries, and the LogicRec queries. 

The main contributions of this work include:
1) We formulate the task of LogicRec, recommendation with both user preferences and their logical requirements, to perform highly personalized recommendations.
2) We propose an initial solution for LogicRec: we employ LQE to deal with the KG incompleteness issue and propose a multi-task knowledge-sharing framework to alleviate the answer set under-exploitation issue.
3) We conduct extensive experiments based on our constructed benchmark datasets to show the significance of the task and the effectiveness of our method.

\section{Related Work}
\textit{Recommender Systems:}
Collaborative filtering (CF) is a commonly used paradigm for recommendation~\cite{he2018nais,he2017neural,wang2019neural}, which is based on user and item interactions.
Also, methods~\cite{rendle2011fast,cheng2016wide,he2017nfm,lian2018xdeepfm} have been designed to model side information, such as item attributes and user profiles.
LogicRec falls into the CF category since we consider user and item indices without side information.
More recently, efforts~\cite{zhang2016collaborative,wang2018ripplenet,wang2019multi,wang2019kgat,wang2021learning,mcclk2022} have been made to incorporate background knowledge from KGs to enrich item and user representations in recommendations (KGRec).
LogicRec is closely related to KGRec methods in that we are also using KGs. However, KGs in LogicRec are not limited to the source of background knowledge, while also backing up the structured logical requirements that users may provide. More specifically, instead of retrieving items based on KG-enriched item and user representations as previous KGRec methods, we additionally model logical requirements over KGs as the queries for item retrieval.

\textit{Logical Query Embedding:}
Logical query embedding (LQE) aims to embed multi-hop queries with logical operations on KGs and entities into the same embedding space, so as to jointly answer multi-hop logical queries in the presence of missing links.
Great efforts have been made to develop LQE systems in recent years, including GQE
\cite{hamilton2018embedding}, Query2Box \cite{ren2019query2box}, HypE \cite{choudhary2021self}, ConE \cite{zhang2021cone}, BetaE \cite{ren2020beta}, \cite{arakelyan2020complex}, LogicE \cite{luus2021logic}, and FuzzQE \cite{chen2022fuzzy}. 
Although we may take the intersection of the user preference and the logical requirement as a whole logical query to directly apply previous LQE methods, the rich knowledge behind the different answer sets of user preference retrieval and logical requirement retrieval remains unexplored. 

\begin{figure}[t!]
	\centering  
	\includegraphics[width=0.38\textwidth]{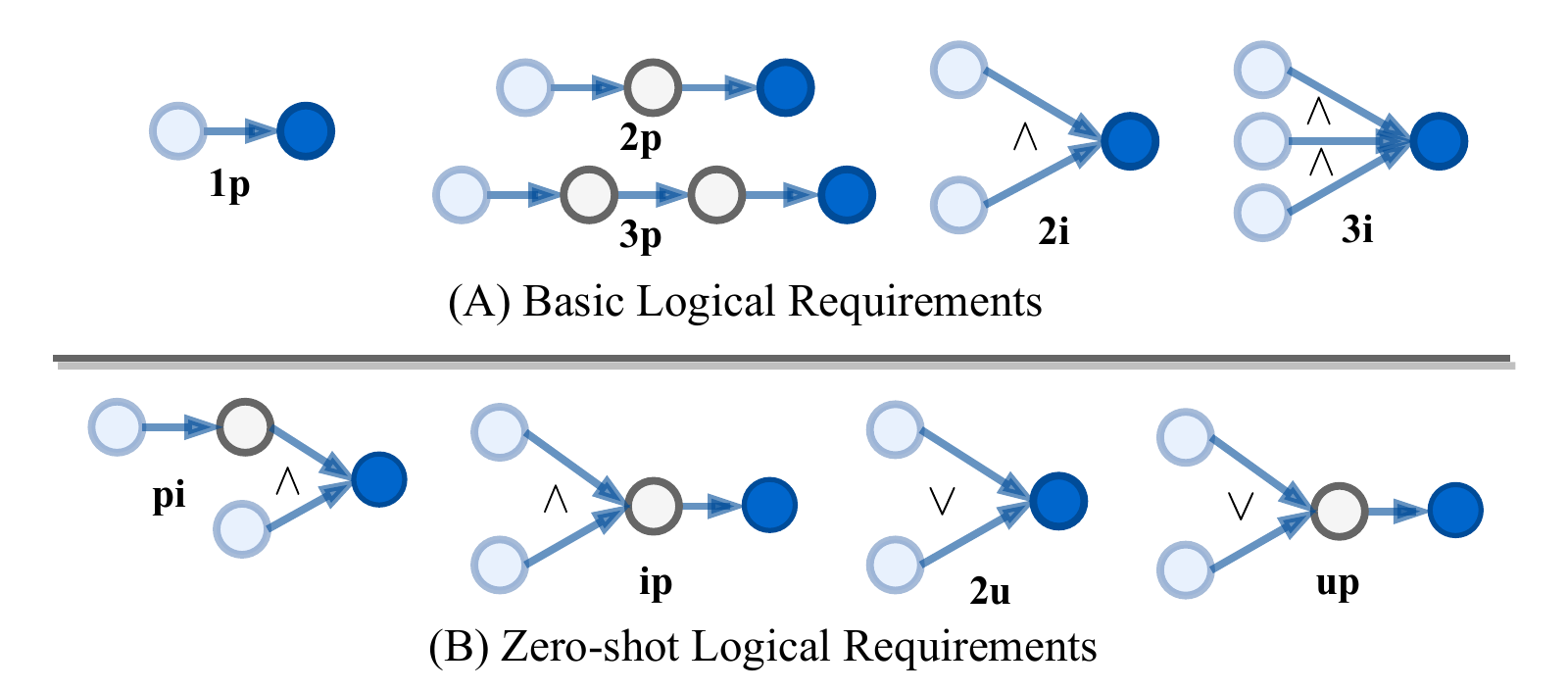}
 \vspace{-0.3cm}
	\caption{The investigated types of logical requirements. (A) Basic logical requirements are used for training and testing, (B) Zero-shot logical requirements are only used for testing the generalization capabilities of our method. \textit{i}, \textit{u}, and \textit{p} denote intersection ($\wedge$), union ($\vee$), and projection, respectively.}
	\label{fig:types}
\vspace{-0.5cm}
\end{figure}

\section{Methodology}
\subsection{LogicRec Formulation}
LogicRec is defined on KGs, where a KG is formulated as $\mathcal{KG} = \{ \langle h, r, t \rangle \} \subseteq \mathcal{E} \times \mathcal{R} \times \mathcal{E}$, $h$, $r$, $t$ denote the head entity, relation, and tail entity in triple $\langle h, r, t \rangle$, respectively, $\mathcal{E}$ and $\mathcal{R}$ refer to the entity set and the relation set in $\mathcal{KG}$. 
Collaborate filtering is defined on a bipartite graph $\mathcal{G} = \{ \langle u, r^*, i \rangle \} \subseteq \mathcal{U} \times \{r^*\} \times \mathcal{I}$, where $u$, $r^*$, $i$ denote the user, the \textit{user$\_$likes} relation, and the item, respectively, $\mathcal{U}$ and $\mathcal{I}$ refer to the user set and the item set in $\mathcal{G}$. In LogicRec, $\mathcal{G}$ is a subset of $\mathcal{KG}$, i.e., $\mathcal{U} \subset \mathcal{E}$, $\{r^*\} \subset \mathcal{R}$, and $\mathcal{I} \subset \mathcal{E}$.
User preference retrieval is to find a set of items as the answer set $\mathcal{A}_u \subset \mathcal{I}$, such that $i \in \mathcal{A}_u$ iff $q_u[i] = True$, where the user preference query-answer pair $q_u[i] = r^*(u, i)$. 
Logical requirement retrieval is to find a set of items as the answer set $\mathcal{A}_l \subset \mathcal{I}$, such that $i \in \mathcal{A}_l$ iff $q_l[i] = True$, where $q_l[i]$ denotes the logical requirement query-answer pair over KGs which can be intersections
$q_l[i] = r_1(e_1, i) \wedge \cdots \wedge r_n(e_n, i)$,
unions
$q_l[i] = r_1(e_1, i) \vee \cdots \vee r_n(e_n, i)$,
projections
$q_l[i] = \exists m_1, \cdots, m_{n-1}: r_1(e_1, m_1) \wedge r_2(m_1, m_2)\wedge \cdots \wedge r_n(m_{n-1}, i)$,
and combinations of them. Note that $m_1, \cdots, m_{n-1} \in \mathcal{E}$ are variable entities, $e_1, \cdots, e_n \in \mathcal{E}$ and $r_1, \cdots, r_n \in \mathcal{R}$ are non-variable entities and relations, respectively. For example, the logical requirement of type \textit{pi} as shown in Figure \ref{fig:types} is the combination of a \underline{p}rojection and an \underline{i}ntersection, i.e., the \textit{pi} query-answer pair $q_l[i] = \exists m: (r_1(e_1, m) \wedge r_2(m, i)) \wedge r_2(e_2, i)$.
LogicRec is to perform recommendations based on both user preferences and users' logical requirements, which can be regarded as finding a set of items as the answer set $\mathcal{A} \subset \mathcal{I}$, such that $i \in \mathcal{A}$ iff $q[i] = q_u[i] \wedge q_l[i] = True$, i.e, $\mathcal{A} = \mathcal{A}_u \cap \mathcal{A}_l$.

\subsection{LogicRec Query Embedding}
Intuitively, the answer set of logical requirements $\mathcal{A}_l$ can be found by traversing the paths linked by triples in KGs. However, KGs are well-known to be severely incomplete with missing true facts~\cite{Tang2022PositiveUnlabeledLW}. Therefore, such a naive method will result in incomplete $\mathcal{A}_l$, thus yielding unsatisfying $\mathcal{A}$.
For example, if $r_1(e_1, i)$ is a missing true fact, $q_l[i] = r_1(e_1, i) \wedge \cdots \wedge r_n(e_n, i) = False$ and $i$ would be a missing answer. 
LQE methods \cite{hamilton2018embedding,ren2019query2box} enable jointly inferring missing facts and answering logical queries by relaxing the absolute answers to approximate answers. We thus resort to LQE methods to model LogicRec queries. 
Since our main focus of this work is to propose an initial solution specifically to LogicRec that can deal with the KG incompleteness issue and the answer set under-exploited issue, rather than develop a novel LQE method, we employ the basic yet effective GQE \cite{hamilton2018embedding} as our base model. We show the process of embedding LogicRec queries below, which are beyond the query types LQE methods were originally used for.

\textit{Logical Requirement Query Embedding:}
The query-answer pair $q_l[i] \in \{True, False\}$ is relaxed to the probability of $q_l[i] = True$. To do so, we embed the queries into the entity embedding space as $\mathbf{q}_l$ following GQE~\cite{hamilton2018embedding}.
The \textbf{projections} are embedded by $\mathbf{q}_l = \mathbf{e_1} + \mathbf{r_1} + \cdots + \mathbf{r_n}$,
where $\mathbf{e}$ and $\mathbf{r}$ denotes the commonly used \cite{he2017neural,he2017nfm} randomly initialized lookup embedding of entities and items of $d$ dimensions.
The \textbf{intersection} of two query embeddings $\mathbf{q_1}$ and
$\mathbf{q_2}$ is resolved by $\mathbf{q}_l = a(\mathbf{q_1} \oplus \mathbf{q_2}; \Omega)_1 * \mathbf{q_1} + a(\mathbf{q_1} \oplus \mathbf{q_2}; \Omega)_2 * \mathbf{q_2}$,
where $\oplus$ denotes matrix concatenation over the last dimension,
$\Omega$ denote s the parameters of $a(\cdot)$, and $a(\cdot)$ is a
two-layer feed-forward network with $Relu$ activation. $a(\cdot)_1$
and $a(\cdot)_2$ represent the first and second attention weights of $d$ dimensions,
respectively. The \textbf{union} of two query embeddings
$\mathbf{q_1}$ and $\mathbf{q_2}$ is resolved by $\mathbf{q}_l = \max(\mathbf{q_1}, \mathbf{q_2})_{-1}$,
where $\max(\cdot)_{-1}$ denotes the max operation over the last
dimension.

\textit{User Preference Query Embedding:} Recall that user preference query-answer pair is formulated as $q_u[i] = r^*(u, i)$, so the user preference query is essentially a \textit{1p} query as shown in Figure \ref{fig:types}. Therefore, we follow the same process of projection operations to embed user preference queries as $\mathbf{q}_u = \mathbf{u} + \mathbf{r^*}$.

\textit{LogicRec Query Embedding:}
LogicRec jointly considers user preferences and their logical requirements, thus LogicRec queries naturally become the intersection of $\mathbf{q}_u$ and $\mathbf{q}_l$: $\mathbf{q} = a(\mathbf{q}_l \oplus \mathbf{q}_u; \Omega)_1 * \mathbf{q}_l + a(\mathbf{q}_l \oplus \mathbf{q}_u; \Omega)_2 * \mathbf{q}_u$,
where the intersection network is the same as in logical requirement query embedding.

\subsection{Multi-task Knowledge Sharing}
The answer set of LogicRec is the intersection of the answer sets of user preference retrieval and logical requirement retrieval, i.e., $\mathcal{A} = \mathcal{A}_u \cap \mathcal{A}_l$. Intuitively, the preference-item and requirement-item pairs in $\mathcal{A}_u$ and $\mathcal{A}_l$ can be used as positive training samples. However, if we simply apply LQE methods, we can only exploit the LogicRec query-answer pairs $q[i]$ for $i \in \mathcal{A}$, whereas the rich knowledge in $q_l[i]$ for $i \in \mathcal{A}_l$ and $q_u[i]$ for $i \in \mathcal{A}_u$ are unexplored. Therefore, we refer to the idea of MMOE~\cite{ma2018modeling} to design a multi-task knowledge-sharing framework to enable knowledge transfer from preference retrieval and requirement retrieval to LogicRec based on the multi-gate mixture of experts. 
\begin{figure}[t!]
	\centering  
	\includegraphics[width=0.45\textwidth]{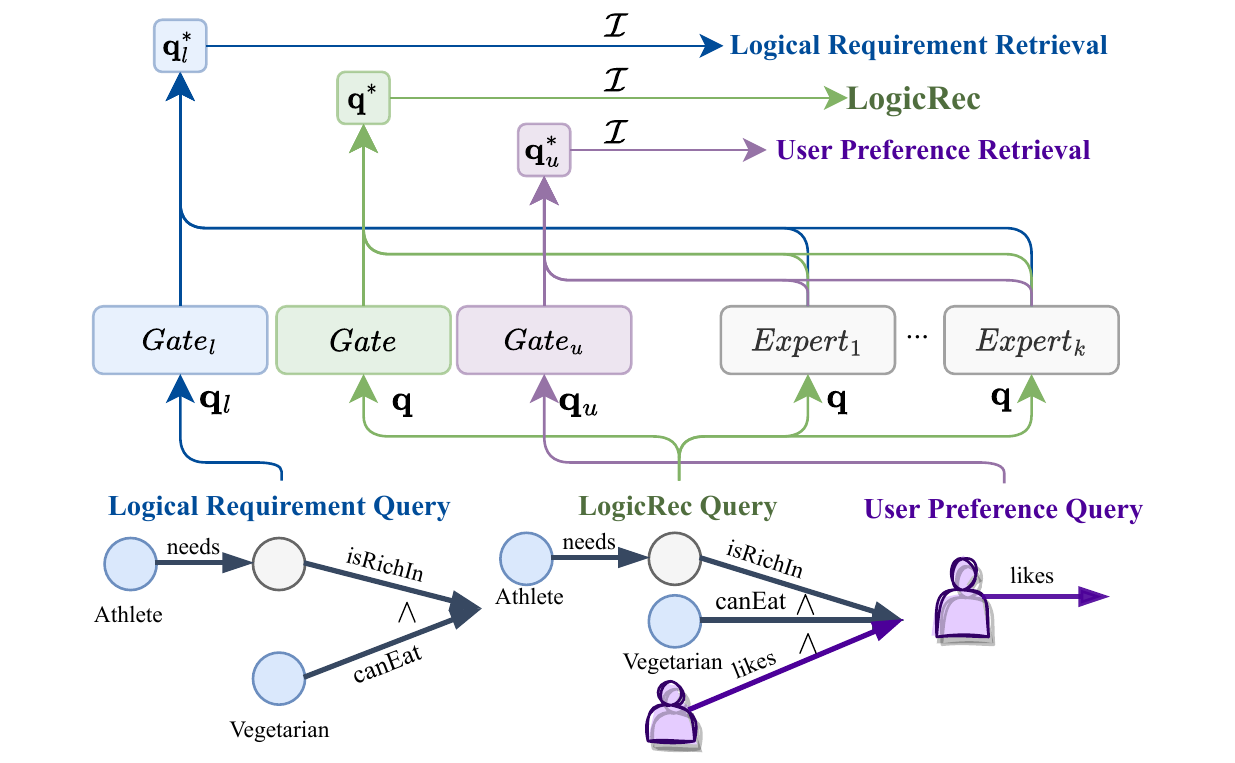}
	\caption{The multi-task knowledge-sharing framework for LogicRec based on the mixture of \textit{k} experts.}
	\label{fig:model}
\end{figure}

Specifically, we first use \textit{k} experts to extract knowledge of $q$ in \textit{k} aspects:
$\mathbf{q}_s = f_s(\mathbf{q}; \theta_{s}), s \in \{1, \cdots k \}$,
where $f_s(\cdot)$ is a single-layer feed-forward network with $Relu$ activation, $\theta_{s} \in \mathbb{R}^{(d + 1) \times d}$ denotes the parameters of $f_s(\cdot)$. We then aggregate the $k$ aspects of knowledge $[\mathbf{q}_1, \cdots, \mathbf{q}_k]$ using gating networks to enable knowledge sharing among multiple tasks as $\mathbf{q}^* = g(\mathbf{q}; \Gamma_g) [\mathbf{q}_1, \cdots, \mathbf{q}_k]^\intercal$, $\mathbf{q}_l^* = g_l(\mathbf{q}_l; \Gamma_{g_l}) [\mathbf{q}_1, \cdots, \mathbf{q}_k]^\intercal$, and $\mathbf{q}_u^* = g_u(\mathbf{q}_u; \Gamma_{g_u}) [\mathbf{q}_1, \cdots, \mathbf{q}_k]^\intercal$,
where $g(\cdot)$, $g_l(\cdot)$, and $g_u(\cdot)$ are single-layer feed-forward networks with softmax activation, $\Gamma_g, \Gamma_{g_l}, \Gamma_{g_u} \in \mathbb{R}^{(d + 1) \times k}$ denote their parameters. As shown in Figure \ref{fig:model}, we perform logical requirement retrieval, user preference retrieval, and LogicRec on top of the query embeddings $\mathbf{q}^*$, $\mathbf{q}_l^*$, and $\mathbf{q}_u^*$ with shared knowledge. Following GQE~\cite{hamilton2018embedding}, the probabilities of $i$ being an answer to the $q$, $q_l$, and $q_u$ are computed as:
$p(q[i]) = \sigma(\gamma - ||\mathbf{q}^* - \mathbf{i}||_1), i \in \mathcal{I}$,
$p(q_l[i]) = \sigma(\gamma - ||\mathbf{q}_l^* - \mathbf{i}||_1), i \in \mathcal{I}$, and
$p(q_u[i]) = \sigma(\gamma - ||\mathbf{q}_u^* - \mathbf{i}||_1), i \in \mathcal{I}$, respectively,
where $\sigma(\cdot)$ is the sigmoid function, and the margin $\gamma$ is a hyperparameter. In this way, the model yields different probabilities for each task out of the same LogicRec query-item pair $q[i]$ with the multi-task knowledge-sharing framework, enabling the exploitation of the answer sets to all tasks, i.e., $\mathcal{A}$, $\mathcal{A}_l$, and $\mathcal{A}_u$.
We apply binary cross-entropy loss for parameter learning and use $p(q[i]), i \in \mathcal{I}$ for inference.

Note that our multi-task knowledge-sharing framework essentially differs from MMOE in two folds. First, MMOE uses the same input for gates and different prediction networks (towers). However, since all our three tasks are inherently LQE tasks, it is more reasonable to use the same tower for all of them. We thus input differently to the gates to distinguish the tasks.
Also, MMOE is used for scenarios where multiple tasks are equally important, while we enable knowledge sharing to facilitate the target task LogicRec.

\begin{table*}[t!]
\small
    \renewcommand\arraystretch{0.8}
  	\renewcommand\tabcolsep{2pt}
  \caption{LogicRec results with various types of logical requirements. We report the averaged \textit{hit@20} over 3 independent runs. }
  \vspace{-0.3cm}
    \begin{tabular}{c||ccccc|cccc|c||ccccc|cccc|c}
    \toprule
          & \multicolumn{10}{c||}{Amazon-book}                                            & \multicolumn{10}{c}{Yelp2018} \\
\cmidrule{2-21}          & \multicolumn{5}{c|}{Basic Logical Requirements} & \multicolumn{4}{c|}{Zero-shot Requirements} &       & \multicolumn{5}{c|}{Basic Logical Requirements} & \multicolumn{4}{c|}{Zero-shot Requirements} &  \\
\cmidrule{2-10}\cmidrule{12-20}          & 1p    & 2p    & 3p    & 2i    & 3i    & ip    & pi    & 2u    & up    & avg   & 1p    & 2p    & 3p    & 2i    & 3i    & ip    & pi    & 2u    & up    & avg \\
    \midrule
    GQE   & 0.072 & 0.056 & 0.057 & 0.172 & 0.256 & 0.168 & 0.071 & 0.082 & 0.081 & 0.113 & 0.050 & 0.029 & 0.042 & 0.130 & 0.176 & 0.111 & 0.028 & 0.035 & 0.031 & 0.070 \\
    Q2B   & 0.034 & 0.044 & 0.050 & 0.158 & 0.292 & 0.156 & 0.054 & 0.059 & 0.071 & 0.102 & 0.027 & 0.043 & 0.035 & 0.106 & 0.211 & 0.143 & 0.023 & 0.032 & 0.023 & 0.071 \\
    BetaE & 0.033 & 0.051 & 0.040 & 0.155 & 0.302 & 0.160 & 0.032 & 0.065 & 0.055 & 0.099 & 0.028 & 0.043 & 0.025 & 0.108 & 0.218 & 0.130 & 0.025 & 0.027 & 0.024 & 0.070 \\
    FuzzQE & \textbf{0.087} & 0.065 & 0.062 & 0.191 & 0.271 & 0.155 & 0.082 & 0.075 & 0.070 & 0.118 & 0.058 & 0.039 & \textbf{0.050} & 0.144 & 0.226 & 0.123 & 0.033 & 0.041 & \textbf{0.034} & 0.083 \\
    \midrule
    \textbf{LogicRec} & 0.086 & \textbf{0.082} & \textbf{0.083} & \textbf{0.228} & \textbf{0.345} & \textbf{0.192} & \textbf{0.101} & \textbf{0.091} & \textbf{0.087} & \textbf{0.144} & \textbf{0.071} & \textbf{0.043} & 0.046 & \textbf{0.176} & \textbf{0.235} & \textbf{0.141} & \textbf{0.040} & \textbf{0.062} & 0.028 & \textbf{0.094} \\
    \bottomrule
    \end{tabular}%
  \label{tab:lqe}%
  \vspace{-0.3cm}
\end{table*}%

\section{Experiments}
We conduct extensive experiments to show the significance of the LogicRec task and the effectiveness of our initial solution for it.
We use the Amazon-book and Yelp2018\footnote{http://jmcauley.ucsd.edu/data/amazon, https://www.yelp.com/dataset} as the raw datasets, which are commonly used for KGRec~\cite{wang2019kgat,mcclk2022}.
We follow the procedures in well-established LQE methods~\cite{ren2019query2box,ren2020beta} to construct LogicRec query-answer pairs. We leave out 5\% of the KG as the known missing facts and we guarantee that the answers to testing LogicRec queries cannot be directly found in the training KG. We apply the commonly used~\cite{ren2019query2box,wang2019kgat} \textit{hit rate} and \textit{ncdg} as evaluation metrics. More details about dataset construction, statistics of datasets, and tuned hyperparameters can be found in our code\footnote{https://github.com/lilv98/LogicRec}. 


\begin{table}[t!]
\small
  \centering
      \renewcommand\arraystretch{0.8}
  \caption{Recommendation performance on Amazon-book.}
  \vspace{-0.3cm}
    \begin{tabular}{c||cccc}
    \toprule
          & hit@10 & hit@20 & ndcg@10 & ndcg@20 \\
    \midrule
    RippleNet~\cite{wang2018ripplenet} &   0.071    & 0.110      &    0.035   & 0.044 \\
    CFKG~\cite{cao2019unifying}  &   0.095    &    \textbf{0.146}   &  0.044     & 0.056  \\
    KGCN~\cite{wang2019knowledge}  &   0.048    &    0.080   &   0.021    & 0.028 \\
    MKR~\cite{wang2019multi}   &   0.042    &    0.070   &    0.018   & 0.025 \\
    \midrule
    GQE   & 0.084 & 0.113 & 0.050 & 0.057 \\
    Shared-Bottom & 0.088 & 0.128 & 0.052 & 0.062 \\
    \midrule
    LogicRec w/o $\mathcal{A}_l$ & 0.072 & 0.097 & 0.048 & 0.048 \\
    LogicRec w/o $\mathcal{A}_u$ & 0.100 & 0.138 & 0.059 & 0.069 \\
    \midrule
    \textbf{LogicRec} & \textbf{0.102} & 0.144 & \textbf{0.061} & \textbf{0.071} \\
    \bottomrule
    \end{tabular}%
  \label{tab:kgrec}%
   \vspace{-0.2cm}
\end{table}%

\textit{More Personalized Recommendations:} We compare our method for LogicRec with representative KGRec methods implemented in a well-recognized toolbox\footnote{https://recbole.io/index.html}.
As the results shown in Table \ref{tab:kgrec}, our method for LogicRec outperforms all KGRec on most of the evaluation metrics. Note that KGRec methods only use KGs to incorporate background information for representation enrichment and still focus on user preference-based recommendation, while we additionally use logical queries on KGs to model users' logical requirements. Such results demonstrate that with the introduction of users' logical requirements, we can capture users' intentions more accurately, and thus provide more personalized recommendations.

\textit{LogicRec Performance:}
We compare our method for LogicRec with representative methods from the three main categories of LQE methods, namely GQE~\cite{hamilton2018embedding} and Q2B~\cite{ren2019query2box} for geometric-based methods, BetaE~\cite{ren2020beta} for distribution-based methods, and FuzzQE~\cite{chen2022fuzzy} for fuzzy logic-based methods. We extend their original implementations
to LogicRec queries for fair comparisons. As shown in Table \ref{tab:lqe}, our proposed method for LogicRec outperforms all baseline methods on most types of logical requirements on both datasets, which clearly demonstrates the effectiveness of our proposed method for LogicRec. Note that directly applied LQE methods can only exploit $\mathcal{A}$, ignoring $\mathcal{A}_l$ and $\mathcal{A}_u$. Such results show that the rich knowledge in $\mathcal{A}_l$ and $\mathcal{A}_u$ is highly helpful for LogicRec and our proposed multi-task knowledge-sharing framework can effectively exploit such knowledge to facilitate LogicRec.

\textit{Generalization Capabilities:}
Since users' logical requirements are not restricted to the forms of the training logical requirements as shown in Figure \ref{fig:types} (A), we also desire satisfiable recommendation performance on unseen (zero-shot) LogicRec queries such as the ones in Figure \ref{fig:types} (B). As shown in Table \ref{tab:lqe}, our method can yield better results on most types of zero-shot LogicRec queries than baseline methods, demonstrating the satisfactory generalization capabilities of our proposed method to unseen LogicRec queries.


\textit{Ablation Study:}
We first compare our method with a basic yet effective multi-task learning method Shared-Bottom \cite{ma2018modeling}. As the results shown in Table \ref{tab:kgrec}, our method can consistently outperform Shared-Bottom on all metrics, demonstrating the effectiveness of our proposed multi-task knowledge-sharing framework for LogicRec. Despite this, Shared-Bottom performs better than GQE, showing that utilizing $\mathcal{A}_l$ and $\mathcal{A}_u$ is helpful for LogicRec and Shared-Bottom can alleviate the answer set under-exploitation issue to some extent. On the other hand, we remove $\mathcal{A}_l$ and $\mathcal{A}_u$ respectively from our proposed method for LogicRec. However, we found that our method performs worse than GQE when $\mathcal{A}_l$ is removed and performs better than GQE when $\mathcal{A}_u$ is removed. We believe the reason is that the logical requirement retrieval is more complicated for the model to learn, while the user preference retrieval is comparatively straightforward. Thus, with the removal of $\mathcal{A}_l$, the model emphasizes too much user preference retrieval with the additional $\mathcal{A}_u$, whereas the logical requirement retrieval is not well-learned. We observe that with $\mathcal{A}$, $\mathcal{A}_l$, and $\mathcal{A}_u$, our method performs the best. Such results demonstrate that $\mathcal{A}_u$ can also contribute to LogicRec with the existence of $\mathcal{A}_l$, and thus shows the effectiveness of our proposed multi-task knowledge-sharing framework in alleviating the answer set under-exploitation issue.

\section{Conclusion}
In this work, we formulated the task of LogicRec, recommendation with both user preferences and their logical requirements, to perform highly personalized recommendations. We proposed an initial solution for LogicRec: we employed LQE to deal with the KG incompleteness issue and proposed a multi-task knowledge-sharing framework to alleviate the answer set under-exploitation issue. Extensive experiments based on our constructed benchmark datasets demonstrated the significance of the LogicRec task and the effectiveness of our proposed method.

\clearpage
\bibliographystyle{ACM-Reference-Format}
\balance
\bibliography{sample-base}

\end{document}